%% file: main.tex
\documentclass{dreamxreport}

\usepackage{pifont}

\input{preamble/commands}

\input{config/brand}
\input{config/report}

\begin{document}

\maketitle
\vfill
\input{sections/teaser}
\clearpage

\tableofcontents
\clearpage

\input{sections/introduction}
\input{sections/data_system}
\input{sections/native_joint_av_generation}
\input{sections/rl}

\input{sections/refiner}
\input{sections/evaluation}
\input{sections/related_work}
\input{sections/conclusion}
\input{sections/acknowledgements}
\input{sections/authors}

\bibliographystyle{dreamxplainnat}
\bibliography{references}

\end{document}

%% file: preamble/commands.tex
\DeclareRobustCommand{\ours}{DreamX-Creator 1.0}
\DeclareRobustCommand{\ourstitle}{\ours}

\newcommand{\refiner}{2K Refiner}

\newcolumntype{L}[1]{>{\raggedright\arraybackslash}m{#1}}
\newcolumntype{C}[1]{>{\centering\arraybackslash}m{#1}}

%% file: config/brand.tex
\DreamXLogo{}

\definecolor{dreamxprimary}{HTML}{0091FF}
\definecolor{dreamxtext}{HTML}{1C2B33}
\definecolor{dreamxbackground}{HTML}{EAF4FB}
\definecolor{dreamxpurple}{HTML}{7366CC}
\definecolor{dreamxcyan}{HTML}{00B4E5}
\definecolor{dreamxlightgray}{HTML}{F5F5F5}
\definecolor{dreamxtablerowalt}{HTML}{EAF4FB}
\definecolor{dreamxheadergray}{HTML}{888888}

\colorlet{amapblue}{dreamxprimary}
\colorlet{amapfg}{dreamxtext}
\colorlet{amapbg}{dreamxbackground}
\colorlet{BrandPurple}{dreamxpurple}
\colorlet{BrandCyan}{dreamxcyan}
\colorlet{LightGray}{dreamxlightgray}
\colorlet{TableRowAlt}{dreamxtablerowalt}
\colorlet{HeaderGray}{dreamxheadergray}

%% file: config/report.tex
\newcommand{\reportteam}{DreamX Team}
\newcommand{\reportdate}{August 2026}

\DreamXRunningTitle{\ours}
\DreamXHeaderLeft{\reportteam}
\DreamXHeaderRight{\reportdate}
\DreamXPDFAuthor{Jiashu Zhu; Yanhao Zheng; Ruitian Tian; Rujing Dang; Shen Zhang; Bingze Song; Jiachen Lei; Ruimin Lin; Jiahong Wu; Xiangxiang Chu}

\title{%
  \texorpdfstring
    {\ourstitle: Democratizing Native Audio-Video Generation at 2K Resolution}
    {DreamX-Creator 1.0}%
}

\DreamXAuthorPlacement{title}
\DreamXTitleByline{}

\author[*]{Jiashu Zhu}
\author[*]{Yanhao Zheng}
\author[*]{Ruitian Tian}
\author[*]{Rujing Dang}
\author{Shen Zhang}
\author{Bingze Song}
\author{Jiachen Lei}
\author{Ruimin Lin}
\author[\dagger]{Jiahong Wu}
\author{Xiangxiang Chu}

\contributioninline{AMAP, Alibaba Group}
\contributioninline[*]{Equal contribution.}\contributioninline[\dagger]{Project Lead.}

\reportabstract{\input{sections/abstract}}


%% file: sections/abstract.tex

Recent video generators often omit audio or synthesize it in a separate stage,
limiting reciprocal modeling of visual dynamics and acoustic events. We present
\ours{}, a compact native joint audio-video generation system centered on a 7B
generator. Conditioned on a first frame and a text prompt, the generator jointly
denoises modality-specialized audio and video streams. The streams are processed
independently in the first half of the network and coupled in the latter half
through Gated Cross-Modal Attention, whose token- and head-wise output gates
modulate each active cross-modal attention-head output. A unified Audio-Video
Data System constructs and filters temporally coherent clips, produces
structured multimodal annotations, and organizes clips into capability-oriented
data pools. Progressive Joint Training comprises two audio-video pre-training
stages followed by High-Quality Finetuning.
Audio-Video Reinforcement Learning further post-trains the generator with
Modality-Aware Multimodal Feedback that routes video-, audio-, and cross-modal
feedback to the corresponding streams. For high-resolution output, our
Autoregressive 1-Step 2K Refinement pipeline adapts a bidirectional multi-step
teacher into an autoregressive multi-step refiner and distills it into a student
requiring one denoising evaluation per temporal chunk. Overall, \ours{} achieves native, synchronized audio-video generation with
performance competitive with state-of-the-art open-source systems. By releasing
our compact 7B generator and 2K Refiner, we seek to democratize native
audio-video generation and provide an accessible foundation for future research
in unified audio-video generative modeling.

%% file: sections/teaser.tex
\begingroup
\centering
\vspace{-10mm}
\includegraphics[width=\linewidth]{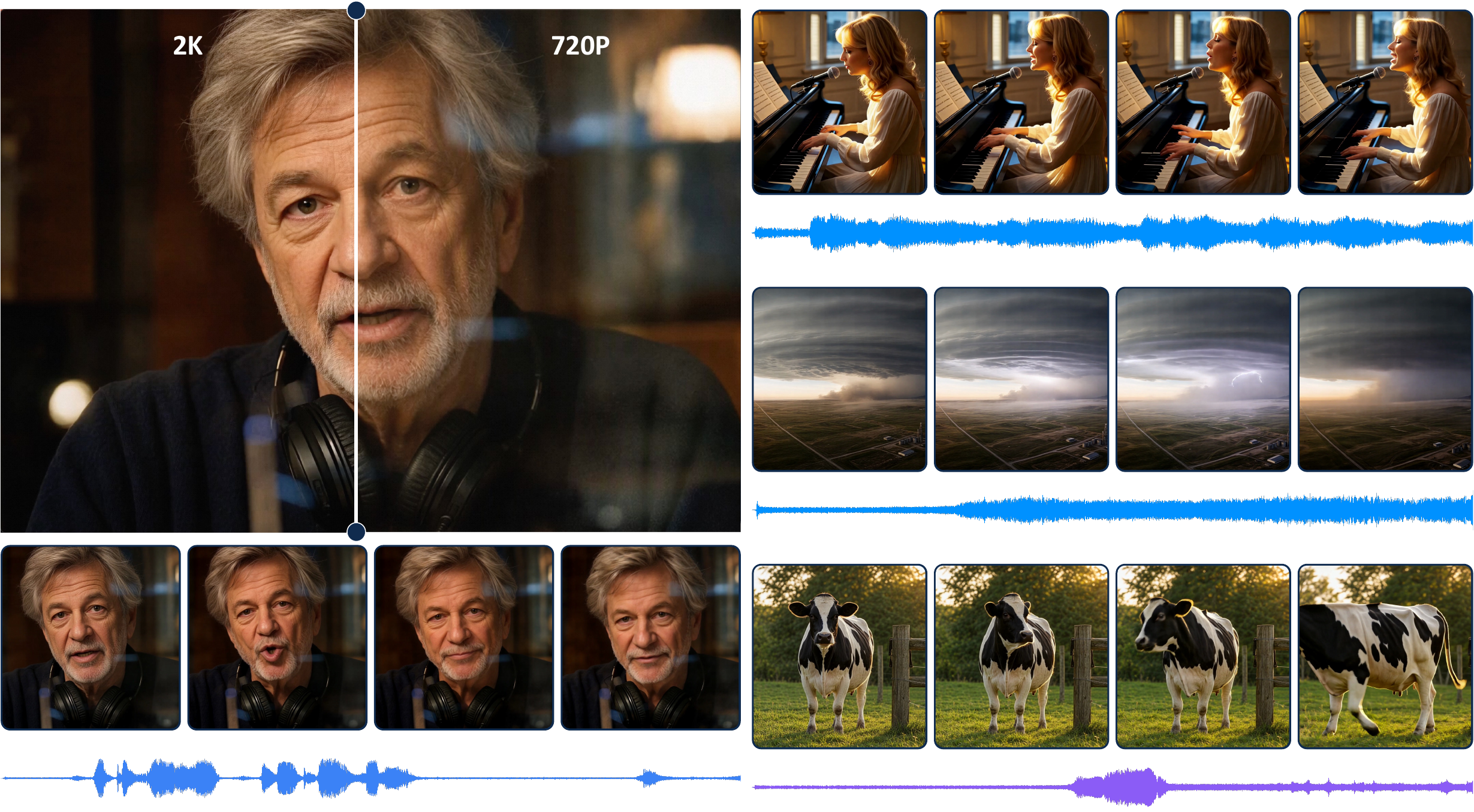}
\captionof{figure}{\textbf{\ours{}.} A native joint generator couples visual motion,
  speech, sound effects, and ambience, followed by multimodal post-training
  and a 2K-oriented video refinement stage.}\label{fig:teaser}
\endgroup

%% file: sections/introduction.tex
\section{Introduction}\label{sec:introduction}

Generative video has advanced rapidly in visual fidelity, motion quality, and
duration, but audio is still often omitted~\citep{team2026dreamx,dreamxteam2026dreamxphi10actionconditionedvideo} or synthesized in a separate stage.
Directional pipelines have achieved strong video-to-audio
generation~\citep{luo2023difffoley,zhang2024foleycrafter,cheng2025mmaudio} and
audio-driven human animation~\citep{tian2024emo,lin2025omnihuman,gao2025wans2v}.
However, these formulations treat one modality as a fixed condition and the
other as the generated output, limiting reciprocal interaction when visual
dynamics and acoustic events are jointly determined. Examples include speech
and mouth motion, visible impacts and their associated sounds, and scene-level
relationships among ambience, music, and camera motion.

Joint audio-video generation instead models both modalities inside the
generative process. Early work coupled audio and video denoisers within joint
diffusion models~\citep{ruan2023mmdiffusion,xing2024seeing,
wang2024avdit,liu2024syncflow}. More recent native generators retain
modality-specialized representations while exchanging information through
dual-stream backbones, expert-based architectures, or unified multimodal
transformers
\citep{low2025ovi,hacohen2026ltx2,wang2025universe,ji2026native,
openmoss2026mova,cheng2026unison,liu2026javisditplusplus,
zhang2026uniavgen}. Together, these systems show that synchronized sound and
video need not be produced as a post-hoc cascade.

This progress exposes four unresolved challenges. \emph{First}, raw audio-video
data require reliable clip segmentation, modality-specific quality filtering,
synchronization assessment, and joint multimodal annotation. Shot transitions
and unrelated background audio can otherwise introduce spurious cross-modal
correlations. \emph{Second}, cross-modal interaction must be introduced without
overwhelming modality-specific representations. Its required strength varies
across layers, attention heads, tokens, and samples, while different content
exhibits different directional dependencies between audio and video.
\emph{Third}, likelihood- or flow-based training does not directly optimize
perceptual quality, prompt adherence, cross-modal semantic consistency, or
fine-grained audio-video synchronization. \emph{Fourth}, joint latent generation
and 2K refinement impose different computational requirements: the former
benefits from a practical latent resolution, whereas the latter must recover
spatial detail while maintaining the generated content, motion, and
audio-synchronized timing.

Model scale and availability remain practical barriers to research on native
audio-video generation. Many existing systems use tens of billions of
parameters, provide only hosted access, or are difficult to reproduce. We
therefore introduce \ours{}, a compact native joint audio-video system centered
on a 7B generator and accompanied by an autoregressive 1-step 2K Refiner. We
publicly release both components to support reproducible evaluation and
downstream adaptation.

The comparison in \cref{tab:av-system-comparison} isolates three system-level
properties relevant to research accessibility and deployment scope:
downloadable model weights, native joint audio-video generation, and an
officially supported output path at 2K resolution or above. Among the listed
systems with a disclosed total backbone size, \ours{} is the smallest model
that combines all three properties.

\begin{table}[H]
  \caption{Capability comparison among representative contemporary video
    generators.}
  \label{tab:av-system-comparison}
  \centering
  \DreamXCompactTableSetup{}
  \begin{tabularx}{\linewidth}{
    @{}X C{0.88in} C{0.82in} C{0.82in} C{0.94in}@{}
  }
    \toprule
    \DreamXTableHeaderRow{}
    \DreamXTableHead{Model} &
    \DreamXTableHead{\shortstack{Backbone params.\\(total / active)}} &
    \DreamXTableHead{\shortstack{Weight\\availability}} &
    \DreamXTableHead{\shortstack{Generation\\modality}} &
    \DreamXTableHead{\shortstack{$\geq$2K\\availability}} \\
    \midrule

    \DreamXTableAltRow{}
    LTX-2.3~\citep{lightricks2026ltx23,lightricks2026ltx23release}
      & 22B
      & Open
      & Joint A/V
      & Local (up to 4K) \\

    MOVA~\citep{openmoss2026mova}
      & 32B / 18B
      & Open
      & Joint A/V
      & Not available \\

    \DreamXTableAltRow{}
    LingBot-Video~\citep{lingbot2026video}
      & 30B / 3B
      & Open
      & Video only
      & Not available \\

    MAGI-2 Preview~\citep{sandai2026magi2}
      & 114B / 6B
      & Open
      & Joint A/V
      & Not available \\

    \DreamXTableAltRow{}
    MiniMax H3~\citep{minimax2026h3}
      & 33B
      & Base only
      & Joint A/V
      & Online only \\

    Seedance 2.5~\citep{bytedance2026seedance25,dreamina2026seedance25}
      & N/D
      & Not released
      & Joint A/V
      & Online only \\

    \DreamXTableAltRow{}
    Kling AI 3.0~\citep{kuaishou2026kling3,kuaishou2026kling4k}
      & N/D
      & Not released
      & Joint A/V
      & Online only \\

    \textbf{\ours{}}
      & \textbf{7B}
      & \textbf{Open}
      & \textbf{Joint A/V}
      & \textbf{Local} \\

    \bottomrule
  \end{tabularx}

  \DreamXTableNote{\raggedright
    Availability is based on publicly documented capabilities as of
    August 27, 2026. ``Open'' denotes downloadable model parameters and
    does not imply an OSI-approved license; ``Base only'' indicates that
    only the base-generation checkpoint is downloadable. In the
    $\geq$2K column, ``Local'' denotes an officially supported local
    generation, refinement, or upscaling path; ``Online only'' denotes
    availability exclusively through an official web service, cloud
    platform, or API; and ``Not available'' indicates that no official
    support was identified. MiniMax H3 releases H3-Base for local 768p
    generation, whereas H3-Context-IR and H3-Regenerate-2K remain
    online-only. LTX-2.3 releases both its generation checkpoints and
    x2 latent spatial upscaler weights, with official support for local
    output up to 4K. Parameter counts refer to the generation backbone;
    active parameters are additionally reported for sparse models.
    N/D: not disclosed; A/V: audio-video.}
\end{table}

At the core of \ours{} is a native joint generator with modality-specialized
audio and video streams. The first half of the network processes the two streams
independently, while the latter half introduces paired audio-to-video (A2V) and
video-to-audio (V2A) paths through Gated Cross-Modal Attention. The paths operate
on positions mapped to a shared temporal coordinate system, with temporal
rotary position encoding applied to their cross-modal queries and keys.
Following the output-gating design studied in gated
attention~\citep{qiu2025gatedattention}, each enabled path uses a token- and
head-wise sigmoid output gate conditioned on both the target hidden state and
the cross-modal attention output. A per-sample direction mask selects the active
path or paths, whereas the gate scales each attention-head output before head
concatenation, output projection, and residual addition.

The complete \ours{} system is supported by a unified Audio-Video Data System.
The base generator is trained with Progressive Joint Training, comprising two
audio-video pre-training stages followed by High-Quality Finetuning. The
internal A2V, V2A, and Joint conditioning configurations are mixed throughout
training and do not define separate inference tasks. Audio-Video Reinforcement
Learning then post-trains the generator using Modality-Aware Multimodal Feedback
that routes video-, audio-, and cross-modal advantages to the corresponding
streams. Finally, \refiner{} adapts a bidirectional multi-step teacher into an
autoregressive multi-step refiner and distills it into an autoregressive 1-step
student. At inference, it enhances the low-resolution video to 2K with one
denoising evaluation per temporal chunk while leaving the audio stream
unchanged.

\begin{figure}[htbp]
  \centering
  \includegraphics[width=\linewidth]{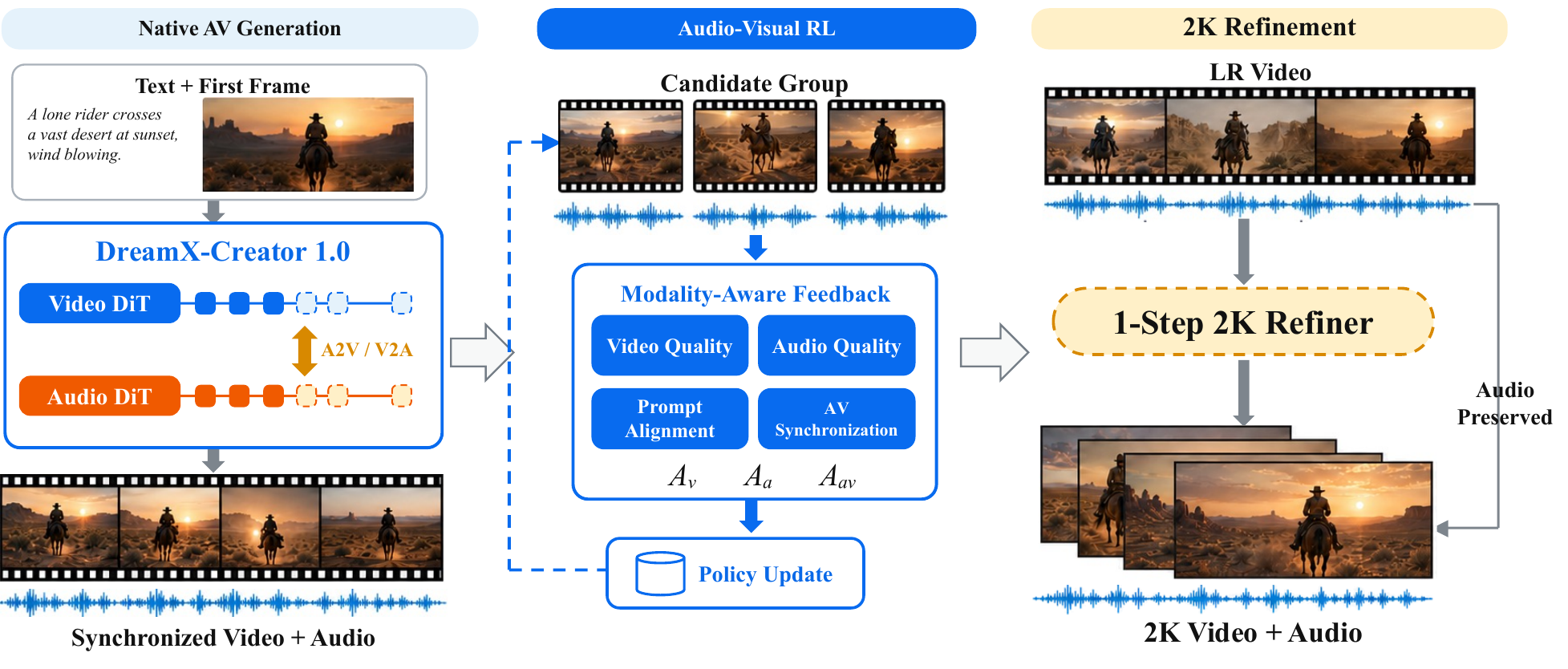}
  \caption{The \ours{} pipeline, comprising native joint audio-video generation,
    Audio-Video Reinforcement Learning post-training, and Autoregressive 1-Step
    2K Refinement.}
  \label{fig:system-overview}
\end{figure}

\begin{maincontributions}
  \item \textbf{Audio-Video Data System.}
    We build a unified pipeline for clip construction and filtering, structured
    multimodal annotation, and capability-oriented data organization to support
    targeted audio-video training.

  \item \textbf{Native Joint Audio-Video Generation.}
    We develop a native joint generator with Gated Cross-Modal Attention and
    Progressive Joint Training, followed by Audio-Video Reinforcement Learning
    with Modality-Aware Multimodal Feedback.

  \item \textbf{Autoregressive 1-Step 2K Refinement.}
    We adapt a bidirectional multi-step teacher into an autoregressive multi-step
    refiner and distill it into a 1-step student that requires one denoising
    evaluation per temporal chunk.

  \item \textbf{A Compact 7B Open-Weight Model.}
    We publicly release our 7B native joint audio-video generator and 2K
    Refiner. The generator is the smallest open-weight native joint model by
    disclosed total backbone size in \cref{tab:av-system-comparison} and
    achieves top-tier benchmark performance.
\end{maincontributions}

%% file: sections/data_system.tex
\section{Audio-Video Data System}\label{sec:data}

Training a general-purpose native audio-video generation model requires a corpus with high perceptual quality, reliable cross-modal correspondence, and diverse audio-visual interactions.
However, raw videos collected from heterogeneous sources are often noisy, weakly aligned, and highly imbalanced in their audio-visual content.
We therefore develop a unified data system that transforms raw videos into training-ready clips through staged filtering for quality and alignment, followed by structured multimodal annotation.
We further organize the curated clips according to their cross-modal dependency patterns, allowing different types of audio-visual data to provide supervision tailored to different training objectives.
We describe each component of the system below.

\begin{figure}[htbp]
  \centering
  \includegraphics[width=\linewidth]{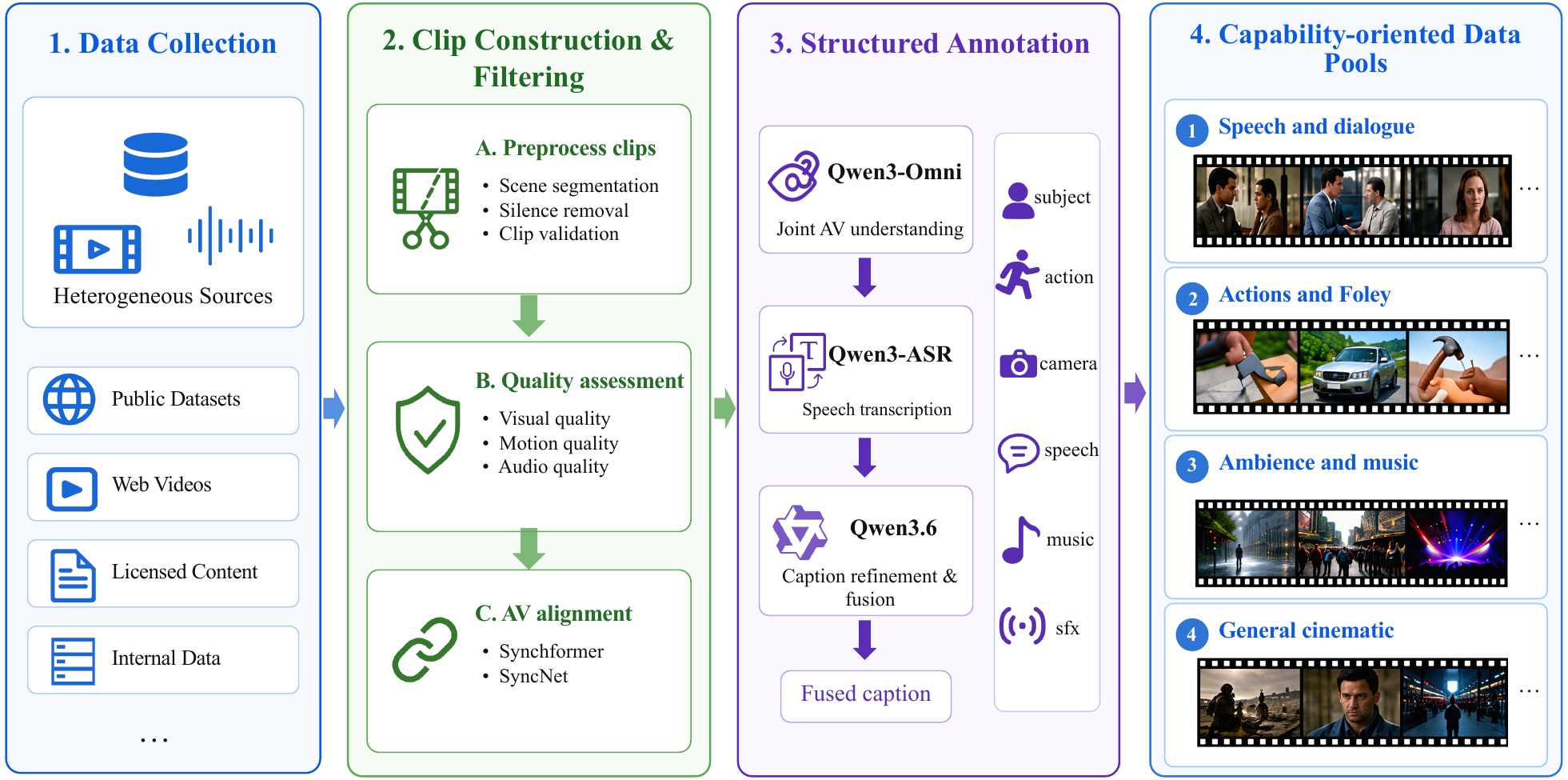}
  \caption{Overview of the data system for \ours{}, including heterogeneous data collection, clip construction and filtering, structured multimodal annotation, and capability-oriented data organization.}
  \label{fig:data-engine}
\end{figure}

\subsection{Data Collection}

We construct our audio-visual corpus by integrating multiple public datasets,
including Koala-36M~\citep{wang2025koala}, VGGSound~\citep{chen2020vggsound},
AudioSet~\citep{gemmeke2017audioset}, OpenHumanVid~\citep{li2024openhumanvid},
SpeakerVid-5M~\citep{zhang2026speakervid}, Action-100M~\citep{chen2026action100m},
and Talker-T2AV~\citep{ye2026talker}, together with internally collected
data. These heterogeneous sources provide complementary coverage across
different content domains and audio-visual interactions. 

\subsection{Data Filtering}

We first preprocess the collected videos into valid audio-visual clips by segmenting them at scene boundaries with PySceneDetect~\citep{Castellano_PySceneDetect}, removing near-silent segments based on audio RMS energy, and discarding invalid or low-resolution samples. As scene-boundary localization may be imperfect, we further trim three frames from both ends of each segment to reduce boundary artifacts and potential cross-shot contamination. We then apply a staged filtering pipeline to assess visual quality, motion, audio quality, and cross-modal synchronization.

\paragraph{Multidimensional quality assessment.}
We use Q-Align to evaluate visual perceptual quality~\citep{wu2023q}, UniMatch-based optical flow to estimate motion magnitude~\citep{xu2023unifying}, and Audiobox Aesthetics to assess audio quality~\citep{tjandra2025audiobox}.

\paragraph{Audio-visual alignment assessment.} We assess temporal alignment at both general and speech-specific levels. Synchformer is used to evaluate synchronization between temporally correlated audio and visual events~\citep{iashin2024synchformer}, while SyncNet is additionally applied to clips with visible speech for fine-grained lip-audio assessment~\citep{chung2016outoftime}. Accordingly, general audio-visual clips are filtered based on Synchformer, whereas visible-speech clips are further required to satisfy the SyncNet-based lip-sync criterion.

\subsection{Data Annotation}

Following filtering, we construct structured multimodal annotations that jointly capture visual content, acoustic events, cross-modal relationships, and spoken text. Our pipeline consists of joint audio-visual understanding followed by caption refinement and fusion.

We first employ Qwen3-Omni-30B-A3B-Instruct to jointly analyze the video and audio streams~\citep{xu2025qwen3}. Unlike pipelines that caption the two modalities independently and merge them afterward~\citep{openmoss2026mova,huang2025jova}, joint multimodal perception allows visual and acoustic evidence to constrain each other, helping reduce hallucinated or cross-modally inconsistent descriptions. The resulting annotation describes subjects, actions, scene context, camera view, speech, music, sound events, and ambience, while organizing audio-visual events in temporal order and distinguishing visually grounded sounds from off-screen audio.

To improve the accuracy of spoken content, we additionally use Qwen3-ASR-1.7B to transcribe speech~\citep{shi2026qwen3}. The multimodal annotation and ASR transcript are then jointly provided to Qwen3.6-27B~\citep{qwen3.6-27b}, which consolidates them into a coherent and fluent caption while preserving the temporal structure and spoken content. The resulting caption is used as the conditioning signal for audio-visual generation.

\subsection{Capability Taxonomy}
\begin{wrapfigure}[15]{r}{0.32\textwidth}
  \centering
  \includegraphics[width=\linewidth]{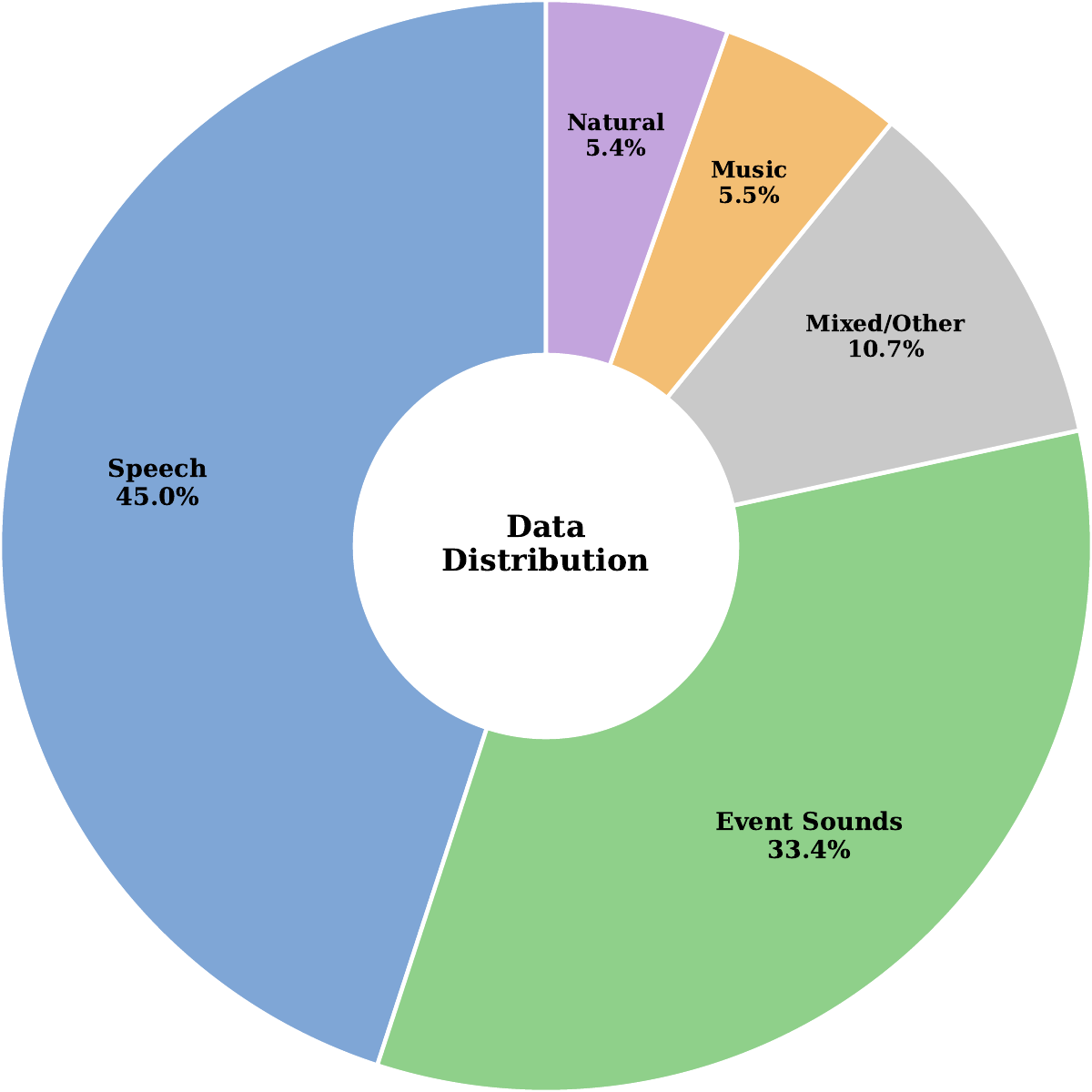}
  \captionsetup{skip=2pt}
  \caption{Content distribution of the training dataset.}
  \label{fig:data-distribution}
\end{wrapfigure}
As shown in~\Cref{tab:data-pools}, we organize the annotated clips into four capability-oriented pools according to their content characteristics and cross-modal dependency patterns. Since different audio-visual content provides distinct forms of cross-modal supervision, we group clips by their dominant supervision signals to support targeted learning of different generation capabilities. Specifically, we use Qwen3.6-27B~\citep{qwen3.6-27b} to infer the dominant cross-modal supervision pattern from each structured audio-visual annotation and assign the clip to the corresponding capability-oriented pool.

\begin{table}
  \caption{Capability-oriented data pools for audio-video generation.}
  \label{tab:data-pools}
  \centering
  \DreamXCompactTableSetup{}
  \begin{tabularx}{\linewidth}{@{}p{1.05in}p{1.55in}XX@{}}
    \toprule
      \DreamXTableHeaderRow{}
      \DreamXTableHead{Pool} &
      \DreamXTableHead{Primary content} &
      \DreamXTableHead{Supervision signal} &
      \DreamXTableHead{Training roles} \\
    \midrule

      \DreamXTableAltRow{}
      Speech and dialogue &
      Visible speech, conversations, and speaker interactions &
      Audio provides fine-grained cues for facial and articulatory dynamics &
      A2V and joint \\

      Actions and Foley &
      Human actions, object interactions, impacts, tools, and vehicles &
      Visible events provide strong cues for temporally aligned sound generation &
      V2A and joint \\

      \DreamXTableAltRow{}
      Ambience and music &
      Environmental ambience, crowds, weather, and background music &
      Visual context provides scene-level cues for sustained acoustic content &
      V2A and joint \\

      General cinematic &
      Mixed scenes without a single dominant audio-visual interaction &
      Mixed cross-modal dependencies provide broad supervision for general audio-visual modeling &
      A2V, V2A, and joint \\
    \bottomrule
  \end{tabularx}
\end{table}



This capability-oriented organization enables us to construct data according to the supervision required by different generation objectives, rather than sampling uniformly from a heterogeneous corpus. By concentrating clips with strong and relevant cross-modal signals, the resulting pools support more targeted capability learning while preserving the diversity needed for general-purpose audio-visual generation.


In addition, we analyze the content distribution of the training data, as shown in Figure~\ref{fig:data-distribution}. Speech accounts for 45.0\% of the data, while event sounds constitute another 33.4\%, including sounds associated with human actions, object interactions, transportation, and other physical events. The remaining samples cover music, natural sounds, and mixed content, providing broad supervision beyond speech-centric scenarios for general audio-visual generation.


%% file: sections/native_joint_av_generation.tex
\FloatBarrier

\section{Native Joint Audio-Video Generation}\label{sec:av-generation}

\subsection{Architecture Overview}

Given a first frame and a text prompt, \ours{} jointly denoises video and audio
latent streams. Each stream retains its own token rate, positional encoding,
and transformer backbone, while receiving text conditioning from a shared text
encoder through a modality-specific conditioning path.

The first half of the network processes the two streams independently. The
latter half adds paired audio-to-video (A2V) and video-to-audio (V2A)
cross-attention paths. A2V uses video queries with audio keys and values; V2A
reverses these roles. When both paths are active, they read the same pre-fusion
states and compute their residual updates in parallel. A direction mask selects
one path for a directional mode or both paths for Joint mode. A2V and V2A are
therefore internal cross-modal paths rather than standalone inference tasks.

Because the streams have different token rates, we map their positions to a
shared temporal coordinate system and apply temporal rotary position encoding
to cross-modal queries and keys. This provides time-aware attention without
resampling either latent sequence or replacing its modality-specific positional
encoding~\citep{low2025ovi,hacohen2026ltx2,openmoss2026mova,ji2026native}.

\paragraph{Training modes.}
Each sample is assigned an A2V, V2A, or Joint mode. The mode specifies the
active cross-modal path, relative corruption ordering, and stop-gradient rule.
Let \(m_{a\rightarrow v},m_{v\rightarrow a}\in\{0,1\}\) denote the direction
masks:

\begin{equation}
  \begin{aligned}
    \mathrm{A2V}:&\quad
      \sigma_v > \sigma_a,
      & (m_{a\rightarrow v},m_{v\rightarrow a}) &= (1,0),\\
    \mathrm{V2A}:&\quad
      \sigma_a > \sigma_v,
      & (m_{a\rightarrow v},m_{v\rightarrow a}) &= (0,1),\\
    \mathrm{Joint}:&\quad
      \sigma_v = \sigma_a,
      & (m_{a\rightarrow v},m_{v\rightarrow a}) &= (1,1).
  \end{aligned}
  \label{eq:directional-sigma}
\end{equation}

A larger \(\sigma\) denotes stronger corruption, so A2V and V2A use a noisier
target stream and a cleaner conditioning stream. Both streams retain their
flow-matching losses. Let \(h^v\) and \(h^a\) denote the pre-fusion hidden
states. The cross-modal updates are

\begin{equation}
  \begin{aligned}
    \Delta h^v
      &=m_{a\rightarrow v}\,
        \operatorname{Attn}\!\left(
          Q(h^v),K(\widehat{h}^{a}),V(\widehat{h}^{a})
        \right),\\
    \Delta h^a
      &=m_{v\rightarrow a}\,
        \operatorname{Attn}\!\left(
          Q(h^a),K(\widehat{h}^{v}),V(\widehat{h}^{v})
        \right),
  \end{aligned}
  \label{eq:directional-stopgrad}
\end{equation}

For A2V, \(\widehat{h}^{a}=\operatorname{sg}(h^a)\); for V2A,
\(\widehat{h}^{v}=\operatorname{sg}(h^v)\); Joint mode uses the original hidden
states in both paths. Stop-gradient is applied before the key and value
projections. It blocks the target-stream loss from updating the conditioning
backbone through cross-modal attention while leaving the attention parameters
trainable; the conditioning backbone is still trained by its own
flow-matching loss.

\begin{figure}[htbp]
  \centering
  \includegraphics[width=\linewidth]{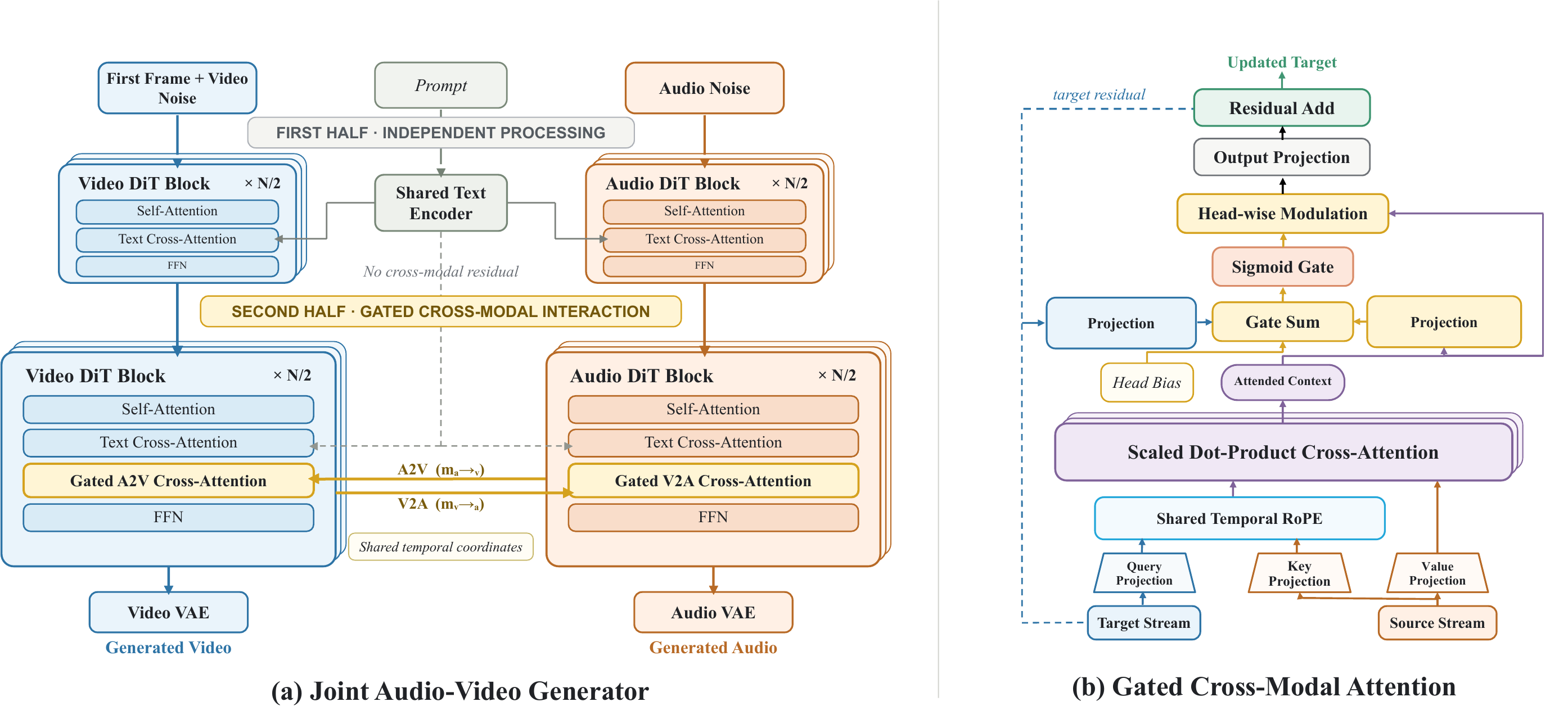}
  \caption{Architecture of the joint audio-video generator. The two streams
    are processed independently in the first half and interact through gated
    A2V and V2A cross-attention in the latter half.}
  \label{fig:av-architecture}
\end{figure}

\subsection{Gated Cross-Modal Attention}\label{sec:gate}

Within each latter-half block, cross-modal attention follows the
modality-specific self-attention and text cross-attention transformations.

More specifically, let \(X\in\mathbb{R}^{L_x\times d_x}\) denote the target
stream and \(Y\in\mathbb{R}^{L_y\times d_y}\) the source stream of a generic
cross-modal path. Each direction uses its own projections to map the two
streams to a common multi-head attention space:

\begin{equation}
  \begin{aligned}
    \bar X &= \operatorname{LN}_{x}(X),
    & \bar Y &= \operatorname{LN}_{y}(Y),\\
    Q &= \operatorname{RMSNorm}(W_Q\bar X),
    & K &= \operatorname{RMSNorm}(W_K\bar Y),
    & V &= W_V\bar Y.
  \end{aligned}
  \label{eq:cross-modal-projection}
\end{equation}

After splitting \(Q\), \(K\), and \(V\) into attention heads, temporal RoPE is
applied using coordinates \(\tau_x\) and \(\tau_y\) in the shared temporal
coordinate system. For head \(h\), the cross-modal attention output is

\begin{equation}
  \begin{aligned}
    \widetilde Q_h
      &=\operatorname{RoPE}_{t}(Q_h,\tau_x),
    & \widetilde K_h
      &=\operatorname{RoPE}_{t}(K_h,\tau_y),\\
    A_h
      &=\operatorname{softmax}\!\left(
        \frac{\widetilde Q_h\widetilde K_h^{\top}}{\sqrt{d_h}}+M_y
      \right),
    & C_h&=A_hV_h,
  \end{aligned}
  \label{eq:cross-modal-attention}
\end{equation}

where \(M_y\) masks padded positions beyond the valid source-stream length.
For A2V, \((X,Y)=(h^v,h^a)\); for V2A,
\((X,Y)=(h^a,h^v)\).

Following the output-gating placement studied in gated
attention~\citep{qiu2025gatedattention}, we apply a token- and head-wise
sigmoid output gate after scaled dot-product attention and before the output
projection. Unlike a query-only gate, our gate depends on both the target token
\(x_i\) and its cross-modal attention output \(C_{i,h}\):

\begin{equation}
  \begin{aligned}
    g_{i,h}
      &=\operatorname{sigmoid}\!\left(
        [W_x\operatorname{LN}(x_i)]_h
        + \mathbf{w}_c^{\top}\operatorname{LN}(C_{i,h})
        + b_h
      \right),\\
    \Delta x_i
      &=W_o\!\left(
        \operatorname{Concat}_{h}
        \left[g_{i,h}\odot C_{i,h}\right]
      \right),\\
    x'_i
      &=x_i+\Delta x_i.
  \end{aligned}
  \label{eq:cross-modal-output-gate}
\end{equation}

Here, \(W_x\) produces one gate logit per head from the target hidden state,
\(\mathbf{w}_c\) maps the normalized head-wise attention output to a scalar,
and \(b_h\) is a head-specific bias. Thus \(g_{i,h}\in(0,1)\) scales each
attention-head output before head concatenation, output projection, and
residual addition.

The output gate and direction mask act at different levels: the gate scales
each active attention-head output, while the direction mask selects whether
the complete A2V or V2A residual is added. The gate is therefore an output
modulation mechanism rather than token routing or path selection.

\subsection{Progressive Joint Training}

Training consists of two audio-video pre-training stages followed by
High-Quality Finetuning. Across all stages, we mix the internal A2V, V2A, and
Joint conditioning configurations defined above.
Pre-training uses a fixed mixture, while High-Quality Finetuning adjusts their
relative emphasis according to the predominant cross-modal dependency of the
training content. These configurations determine the conditioning paths used
during training and do not define separate inference tasks.

\paragraph{Stage 1: LoRA-Based AV Pre-training.}
We initialize from modality-specific video and audio backbones and train on
broad-coverage paired data. The first half of both backbones remains frozen,
rank-256 LoRA adapters are applied to the latter half, and the cross-modal
attention modules and output gates are optimized directly.

\paragraph{Stage 2: Full-Parameter AV Pre-training.}
The Stage 1 LoRA weights are merged into their respective backbones. Training
then continues on broad-coverage paired data with both diffusion-transformer
backbones and all cross-modal modules jointly optimized.

\paragraph{Stage 3: High-Quality Finetuning.}
We construct a curated finetuning subset by filtering candidate clips using
audio-video synchronization metrics, modality-specific aesthetic scores for
both video and audio, spatial resolution, and clip duration. To obtain
temporally coherent clips, we further apply OmniShotCut
\citep{wang2026omnishotcut} for fine-grained shot-boundary detection and
transition labeling, and discard candidate segments that contain or span a
detected transition. We then finetune the full-parameter model on the retained
pairs. The first finetuning run inherits the group-wise learning rates from
Stage 2; subsequent runs progressively reduce them.

\subsection{Optimization Details}

We optimize the video and audio streams using flow matching
\citep{lipman2023flowmatching}. For modality \(m\in\{v,a\}\), let \(z_0^m\)
denote a clean latent, \(\epsilon^m\sim\mathcal{N}(0,I)\) independently sampled
Gaussian noise, and \(\sigma_m\) its modality-specific noise level. The
interpolated latent and target velocity are

\begin{equation}
  z_{\sigma_m}^{m}=(1-\sigma_m)z_0^m+\sigma_m\epsilon^m,
  \qquad
  u_m^*=\epsilon^m-z_0^m.
  \label{eq:av-flow}
\end{equation}

Let \(N_m\) and \(d_m\) denote the number of valid latent tokens and their
feature dimension for modality \(m\), respectively. We define the token- and
feature-normalized flow-matching objective as

\begin{equation}
  \mathcal{L}_{\mathrm{FM}}^m
  =\frac{1}{N_m d_m}\sum_{j=1}^{N_m}
    \left\|\widehat{u}_{\theta,j}^m-u_{m,j}^*\right\|_2^2,
  \qquad
  \mathcal{L}_{\mathrm{AV}}
  =\lambda_v\mathcal{L}_{\mathrm{FM}}^v
  +\lambda_a\mathcal{L}_{\mathrm{FM}}^a,
  \label{eq:av-loss}
\end{equation}

where \(\lambda_v\) and \(\lambda_a\) balance the two modalities. The two
streams use independently sampled Gaussian noise tensors. A base timestep is
drawn from a discrete 1,000-timestep flow schedule with a shift factor of 5.0,
and the modality-specific noise levels follow the directional ordering in
\cref{eq:directional-sigma}. The two audio-video pre-training stages use
\(\lambda_v=\lambda_a=0.5\), whereas High-Quality Finetuning uses
\(\lambda_v=0.5\) and \(\lambda_a=0.1\).

\paragraph{Optimizer.}
In Stage 1, the latter-half LoRA parameters use a learning rate of
\(1\times10^{-4}\), while the cross-modal attention and output-gate parameters
use \(2\times10^{-5}\). LoRA uses rank 256, a scaling factor of 128, and zero
dropout. After merging the adapters, Stage 2 updates the video and audio
backbone parameters with \(1\times10^{-5}\) and the cross-modal modules with
\(2\times10^{-5}\). High-Quality Finetuning begins with the same group-wise
rates and progressively reduces them across successive runs.

All stages use AdamW with \((\beta_1,\beta_2)=(0.9,0.99)\) and
\(\epsilon=10^{-10}\). Weight decay is zero for the Stage 1 LoRA and
cross-modal parameter groups. As specified in our training recipe, Stage 2
uses \(10^{-5}\) for the backbone parameters and zero for the cross-modal
modules, while High-Quality Finetuning uses zero weight decay for both groups.
The two pre-training stages use a 200-step linear warm-up followed by constant
group-wise learning rates. Training uses bfloat16 mixed precision and clips the
gradient norm at 1.0. The complete training pipeline uses a comparatively
modest GPU-day budget.

\FloatBarrier

%% file: sections/rl.tex
\section{Audio-Video Reinforcement Learning}\label{sec:av-rl}

\subsection{Overview}
The native joint generator described in Section~3 is trained with audio and video flow-matching objectives. This training enables the model to capture the joint audio-video distribution and establish bidirectional interaction between the two modality streams. However, minimizing these flow-matching objectives does not directly optimize the properties most relevant to human perception, such as visual and acoustic fidelity, prompt adherence, cross-modal semantic consistency, and precise temporal correspondence between visible events and their associated sounds.

We therefore introduce an audio-visual reinforcement-learning stage to post-train DreamX-Creator 1.0 with multimodal feedback. This stage improves visual and audio quality, semantic consistency, and temporal synchronization while preserving the capabilities and diversity of the base model. Both streams remain within the joint generator, improving their coordination without an additional cascaded generation stage.

A central challenge is that an audio-video sample is evaluated along multiple dimensions that may not improve together. A candidate can contain high-quality video but weak audio, or plausible content in both modalities but incorrect event timing. Consequently, we retain decomposed feedback signals rather than immediately reducing all criteria to a single global score. This design follows the modality-aware optimization principle explored by OmniNFT \citep{zhang2026omninftmodalitywiseomnidiffusion}, while adapting it to the bidirectional architecture and context-aware interaction modules of DreamX-Creator 1.0.

\subsection{Modality-Aware Multimodal Feedback}
A single global reward is insufficient for joint audio-video post-training because the quality of the two modalities may vary independently. A candidate with strong visual quality may contain weak audio, while a perceptually plausible pair may still exhibit incorrect semantic or temporal correspondence. We therefore decompose the feedback into video-specific, audio-specific, and cross-modal components. Let $A_v$, $A_a$, and $A_{av}$ denote their normalized advantages. The supervision routed to the two generation streams is

\begin{equation}
    \widetilde{A}_v = A_v + A_{av},
    \qquad
    \widetilde{A}_a = A_a + A_{av}.
    \label{eq:modality_advantage_routing}
\end{equation}

The modality-specific terms improve the corresponding video or audio stream, whereas the shared cross-modal term jointly supervises both streams and their bidirectional interaction modules. This separation prevents an improvement in one modality from masking degradation in the other and preserves synchronization as a shared optimization objective.

Audio-visual synchronization is mainly determined by a small number of regions, such as visible mouths and sound-producing objects. We use the video-to-audio responses from the selected interaction blocks to estimate the relevance of each video token. To account for response-scale differences across frames, the scores are normalized within each frame and converted into token weights:

\begin{equation}
w_i^{(e)}
=
1+\alpha_e\,
\operatorname{sigmoid}
\left(
\frac{s_i-\mu_{f(i)}}{\sigma_{f(i)}+\epsilon}
\right).
\label{eq:progressive_token_weight}
\end{equation}

Here, $s_i$ is the cross-modal relevance score of video token $i$, and $f(i)$ denotes the frame containing that token. The terms $\mu_{f(i)}$ and $\sigma_{f(i)}$ are the mean and standard deviation of the relevance scores within the same frame, while $\epsilon$ ensures numerical stability. The resulting weight $w_i^{(e)}$ determines the contribution of token $i$ at training step $e$. The coefficient $\alpha_e$ is gradually increased from zero to a predefined maximum during warmup. Thus, training begins with nearly uniform token weights and progressively focuses on regions that contribute more strongly to audio-visual synchronization.

We also apply depth-dependent gradient scaling to the audio-to-video pathway. Gradients entering the audio stream are attenuated more strongly in shallow blocks and progressively restored in deeper interaction blocks. Together, regional weighting and directional gradient scaling improve audiovisual alignment while preserving the pretrained modality-specific representations.

\begin{figure}[htbp]
  \centering
  \includegraphics[width=\linewidth]{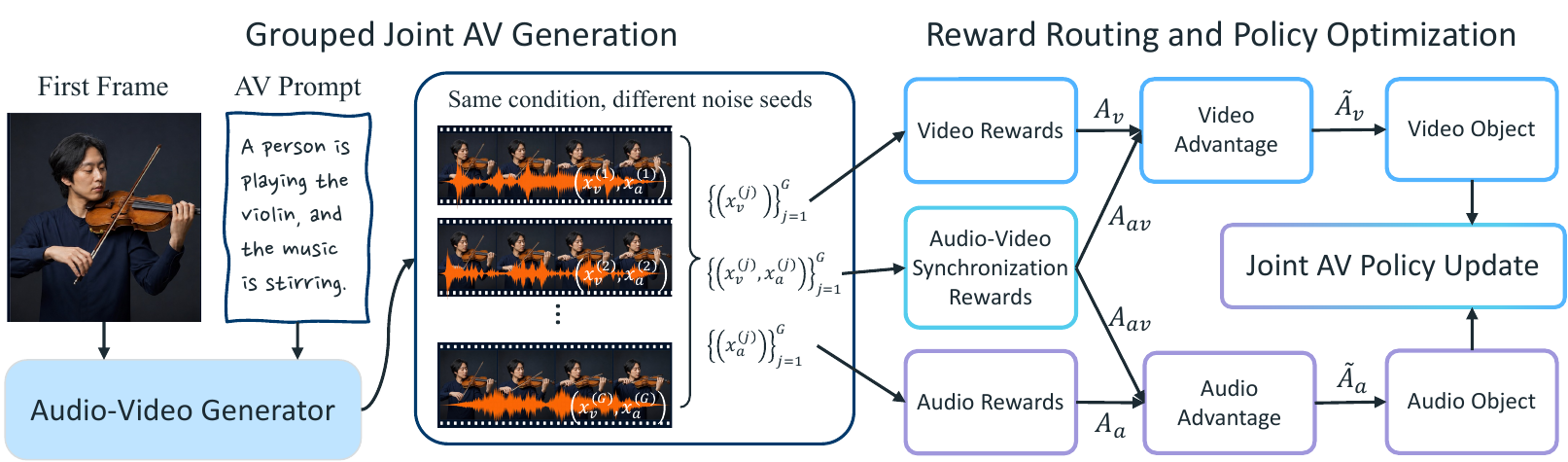}
  \caption{Modality-aware reinforcement learning for joint audio-video generation. Video- and audio-specific rewards are routed to their corresponding streams, while the shared audio-visual synchronization reward jointly optimizes both streams and their cross-modal interactions.}\label{fig:gated-block}
\end{figure}

\subsection{Training}

We initialize the reinforcement-learning policy from the pretrained DreamX-Creator 1.0 base model. The training set contains 1,000 first-frame--prompt pairs, where the first frame specifies the initial visual appearance and scene structure, and the prompt describes the expected visual dynamics and accompanying audio. For each condition $(I,c)$, the current policy generates a group of $G$ joint audio-video candidates:

\begin{equation}
\left\{
\left(x_v^{(j)},x_a^{(j)}\right)
\right\}_{j=1}^{G}
\sim
\pi_{\theta_{\mathrm{old}}}(\cdot \mid I,c),
\label{eq:av_rl_sampling}
\end{equation}

where $\pi_{\theta_{\mathrm{old}}}$ denotes the rollout policy, while $x_v^{(j)}$ and $x_a^{(j)}$ denote the generated video and audio components of the $j$-th candidate, respectively. Each candidate is evaluated independently using video-quality, audio-quality, prompt-consistency, and audio-visual synchronization rewards. The rewards are normalized within candidates generated from the same condition to obtain relative video, audio, and cross-modal advantages.

The resulting advantages are routed according to their corresponding modalities. Video-specific feedback supervises the video stream, audio-specific feedback supervises the audio stream, and cross-modal feedback is shared by both streams and their bidirectional interaction modules. Synchronization-relevant regions identified by deeper cross-modal responses are assigned greater importance, while gradient propagation through shallow modality-specialized blocks is conservatively controlled to protect the visual and acoustic priors learned during pretraining. The overall training objective is written as

\begin{equation}
    \mathcal{L}_{\mathrm{RL}}
    =
    \lambda_v
    \mathcal{L}_v\!\left(A_v + \lambda_{av}^{v}A_{av}\right)
    +
    \lambda_a
    \mathcal{L}_a\!\left(A_a + \lambda_{av}^{a}A_{av}\right)
    +
    \lambda_{\mathrm{reg}}
    \mathcal{L}_{\mathrm{reg}}\!
    \left(\pi_{\theta},\pi_{\mathrm{base}}\right),
    \label{eq:av_rl_objective}
\end{equation}

where $A_v$, $A_a$, and $A_{av}$ denote the group-normalized video, audio, and cross-modal advantages, respectively. The regularization term constrains the updated policy toward the pretrained base model to preserve generation diversity and prevent reward over-optimization.

Training alternates between grouped candidate generation, multimodal reward evaluation, and policy optimization. The updated policy is periodically used to refresh the rollout model, forming an online generation--evaluation--update loop. The post-trained model retains the same first-frame-conditioned joint audio-video generation interface as the base model.

%% file: sections/refiner.tex
\section{Autoregressive 1-Step 2K Refinement}
\label{sec:refiner}

High-resolution video generation requires both coherent motion and rich spatial
detail. Directly generating 2K videos with the joint audio-visual model is
computationally expensive, while operating at a tractable resolution leaves the
output short of fine textures and sharp boundaries. We therefore introduce
\refiner{}, an autoregressive 1-step 2K refinement stage. The joint generator
first produces a coherent low-resolution (LR) video, and \refiner{} enhances it
to 2K resolution while preserving the generated content, motion, and
audio-synchronized timing.

At inference, \refiner{} refines the video sequentially over temporal chunks.
For each chunk, the model performs a single denoising evaluation conditioned on
the LR video information and the previously refined high-resolution (HR)
chunks. This design avoids the cost of multi-step diffusion sampling at 2K
resolution and makes refinement scalable to long videos. The audio stream is
not modified during this process.

Our training pipeline consists of three stages. We first train a bidirectional
multi-step diffusion teacher for high-quality video refinement. We then adapt
it into an autoregressive multi-step refiner, which serves as the intermediate
model for 1-step distillation. Finally, we distill the autoregressive refiner
into a 1-step student with DMD. The deployed \refiner{} is this final
autoregressive 1-step student.

\begin{figure}[htbp]
  \centering
  \includegraphics[width=0.6\linewidth]{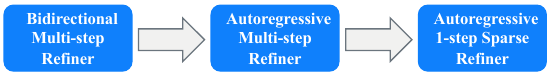}
  \caption{Training pipeline of \refiner{}. A bidirectional multi-step
  refinement teacher is adapted into an autoregressive multi-step refiner and
  then distilled into an autoregressive 1-step 2K student. The final model
  performs one denoising evaluation per temporal chunk refinement.}
  \label{fig:refiner}
\end{figure}

\subsection{Bidirectional Refiner}
\label{sec:refiner_teacher}

We begin with a bidirectional diffusion-transformer teacher trained for video
refinement. The teacher operates in the latent space and is conditioned on the
LR input video. Its bidirectional spatiotemporal attention allows it to exploit
both past and future frames, providing a strong refinement prior for recovering
temporally consistent details.

The teacher is trained with flow matching~\citep{lipman2023flowmatching}. For
each HR training video, we sample a noise level and corrupt the clean HR latent
toward Gaussian noise. The teacher is then optimized to predict the
corresponding velocity field conditioned on the LR video. This objective trains
the model to progressively transform noisy HR latents into clean refined
latents while remaining faithful to the LR input.

\paragraph{Degradation curriculum.}
Training pairs are synthesized by degrading HR videos, following real-world VSR
practice~\citep{chan2022realbasicvsr}. We start from mild degradations to
encourage structure preservation, and progressively introduce stronger blur,
noise, compression artifacts, resampling artifacts, geometric distortion, and
temporally correlated corruptions. The last category is important because the
inputs to \refiner{} are generated videos rather than raw camera captures.
Generated videos often contain flickering textures, local motion jitter, and
unstable fine structures. Including such corruptions during training helps the
teacher learn temporal repair rather than merely spatial sharpening.

\subsection{Autoregressive Refiner}
\label{sec:refiner_ar}

The bidirectional teacher produces high-quality refinements, but running a
multi-step bidirectional model at 2K resolution is too costly. We therefore
factorize HR refinement autoregressively over temporal chunks. Instead of
denoising the entire HR video jointly, the model refines one chunk at a time,
using the LR video information and the previously refined HR context.

This causal factorization reduces high-resolution computation while preserving
temporal continuity through the autoregressive context. It also matches the
intended deployment setting: during inference, each new chunk is refined after
the previous chunks have already been produced.

We convert the
bidirectional teacher into an autoregressive multi-step refiner by conducting teacher forcing: the current chunk is denoised conditioned on the
LR video and ground-truth past HR chunks. The resulting autoregressive refiner
retains the refinement quality of the bidirectional teacher while matching the
temporal structure of the final student.

\subsection{DMD Distillation}
\label{sec:refiner_dmd}

The autoregressive multi-step refiner is still too slow for deployment, since
each chunk requires multiple denoising evaluations. We distill it into a
1-step student using Distribution Matching Distillation
(DMD)~\citep{yin2024onestep}, following the DMD2 formulation
\citep{yin2024dmd2}. DMD matches the output distribution of the student to the
refinement distribution of the multi-step refiner, rather than forcing the
student to imitate intermediate denoising states.

At each autoregressive step, the 1-step student predicts the current HR chunk
from Gaussian noise, the LR video, and its own previously refined chunks. Since
the student uses its own outputs as temporal context at inference, training it
only with ground-truth past chunks would create exposure bias. Small refinement
errors could then accumulate across chunks, leading to flicker, over-sharpening,
or temporal drift.
We therefore train the student under self-rollout. During distillation,
complete videos are generated by the student itself, and the DMD objective is
applied to these generated rollouts. This aligns the training distribution with
the inference distribution and encourages the student to correct errors induced
by its own previous refinements.

The distilled objective combines distribution matching with pixel-space
supervision.  The difference between real-score and fake-score provides the distribution-matching
signal that pushes the student toward the teacher-defined refinement
distribution. In addition, we decode the generated latents with a frozen VAE
decoder and apply perceptual and reconstruction losses against the HR ground
truth. We use DISTS as the perceptual loss and an $\ell_2$ reconstruction loss
for pixel-level consistency, with both loss weights set to one in all
experiments.
The LR video condition keeps the refinement faithful to the generated content,
while distribution matching supplies realistic high-frequency detail. As a
result, \refiner{} performs efficient 1-step 2K refinement without changing the
motion or audio-aligned timing produced by the joint generator.

%% file: sections/evaluation.tex
\section{Evaluation}\label{sec:evaluation}



\subsection{Settings}
While benchmarks such as VMBench~\citep{huang2023vbenchcomprehensivebenchmarksuite, ling2025vmbenchbenchmarkperceptionalignedvideo} effectively evaluate general video generation quality—measuring core attributes like motion and temporal consistency—they are not designed to assess audio quality or audio-visual alignment. Consider this, we evaluate our model on Verse-Bench~\citep{wang2025universe}, which consists of three evaluation sets covering different audio-video generation scenarios.
Set 1 and Set 2 cover diverse general audio-visual events, whereas Set 3 focuses on speech-centric scenarios with visible speakers. Since Verse-Bench provides separate video and audio descriptions for each sample, we use Qwen3.6-27B~\citep{qwen3.6-27b} to merge them into a single unified prompt that jointly describes the visual content and the corresponding acoustic events. The unified
prompts are then used as the text conditions for joint audio-video generation.



\subsection{Quantitative Evaluation}

We evaluate the generated audio-video samples from four complementary aspects:

\begin{itemize}
  \item \textbf{Video quality.}
    We report the video-quality score (VQ), which jointly considers perceptual quality and identity consistency. Perceptual quality is evaluated using Aesthetic Predictor~\citep{aestheticpredictorv25}, MUSIQ~\citep{ke2021musiq}, and MANIQA~\citep{yang2022maniqa}, while identity consistency is measured using DINOv3~\citep{simeoni2025dinov3}.

  \item \textbf{Audio quality.}
  We adopt Audiobox Aesthetics~\citep{tjandra2025audiobox} and report its four dimensions: Content Enjoyment (CE), Content Usefulness (CU), Production Complexity (PC), and Production Quality (PQ), which characterize complementary aspects of generated audio quality.

  \item \textbf{Speech generation.}
  For the speech-focused Set~3, we report Word Error Rate (WER) to evaluate the correctness of generated speech content and SyncNet-based LSE-C~\citep{chung2016outoftime} to measure lip-audio synchronization.

  \item \textbf{Audio-visual alignment.}
  We evaluate both semantic and temporal correspondence between the two modalities. ImageBind similarity (IB)~\citep{girdhar2023imagebind} measures audio-visual semantic consistency, while the Synchformer-based DeSync score~\citep{iashin2024synchformer} measures temporal synchronization.

\end{itemize}

To distinguish comparisons with the five established research baselines from
those with substantially larger recent open-weight systems, we report the two
groups separately. Table~\ref{tab:main-quantitative-comparison} compares
\ours{} with NAVA~\citep{ji2026native}, UniAVGen~\citep{zhang2026uniavgen}, Ovi~\citep{low2025ovi}, and the two DaVinci-MagiHuman variants~\citep{chern2026speed},
whereas Table~\ref{tab:large-open-model-comparison} isolates the comparison
with LTX-2.3~\citep{lightricks2026ltx23,lightricks2026ltx23release} and MiniMax-H3~\citep{minimax2026h3}.

\begin{table}[htbp]
  \caption{Preliminary quantitative comparison with NAVA, UniAVGen, Ovi, and
    the DaVinci-MagiHuman variants. Metrics are grouped by evaluation target,
    and arrows indicate the preferred direction for each metric. Bold and
    underlined values denote the best and second-best results within this
    table, respectively.}
  \label{tab:main-quantitative-comparison}
  \centering
  \DreamXCompactTableSetup{}
  \setlength{\tabcolsep}{1.5pt}
  \begin{tabularx}{\linewidth}{@{}>{\raggedright\arraybackslash}p{1.55in}
      >{\centering\arraybackslash}p{0.40in}
      *{9}{>{\centering\arraybackslash}X}@{}}
    \toprule
    \DreamXTableHeaderRow{}
    \rule{0pt}{2.8ex} & &
    \multicolumn{1}{c}{\DreamXTableHead{Video}} &
    \multicolumn{4}{c}{\DreamXTableHead{Audio Aesthetics}} &
    \multicolumn{1}{c}{\DreamXTableHead{ASR}} &
    \multicolumn{1}{c}{\DreamXTableHead{Lip Sync}} &
    \multicolumn{2}{c}{\DreamXTableHead{AV-Align}} \\[-0.5pt]
    \arrayrulecolor{white}\cline{3-11}\arrayrulecolor{black}
    \cellcolor{dreamxprimary}\rule[-0.8ex]{0pt}{3.2ex}\multirow{-2}{*}{\DreamXTableHead{Model}} &
    \cellcolor{dreamxprimary}\multirow{-2}{*}{\DreamXTableHead{Params}} &
    \cellcolor{dreamxprimary}\DreamXTableHead{VQ$\uparrow$} &
    \cellcolor{dreamxprimary}\DreamXTableHead{CE$\uparrow$} &
    \cellcolor{dreamxprimary}\DreamXTableHead{CU$\uparrow$} &
    \cellcolor{dreamxprimary}\DreamXTableHead{PC$\downarrow$} &
    \cellcolor{dreamxprimary}\DreamXTableHead{PQ$\uparrow$} &
    \cellcolor{dreamxprimary}\DreamXTableHead{WER$\downarrow$} &
    \cellcolor{dreamxprimary}\DreamXTableHead{LSE-C$\uparrow$} &
    \cellcolor{dreamxprimary}\DreamXTableHead{IB$\uparrow$} &
    \multicolumn{1}{>{\columncolor{dreamxprimary}[\tabcolsep][0pt]}c@{}}{\DreamXTableHead{DeSync$\downarrow$}} \\
    \midrule
    \DreamXTableAltRow{}
    NAVA & 6.3B & 0.6116 & 4.7695 & 6.0053 & 2.3027 & 6.3779 & 0.1574 & 7.7261 & {\bfseries 0.2853} & 0.2342 \\
    UniAVGen & 7.1B & \underline{0.6644} & 4.5938 & 5.8605 & \underline{2.1534} & {\bfseries 6.7643} & 0.1661 & 4.9556 & 0.1556 & 0.5371 \\
    \DreamXTableAltRow{}
    Ovi & 10B & 0.6542 & 4.7756 & 5.9832 & {\bfseries 2.1280} & 6.1427 & {\bfseries 0.1053} & 7.3095 & 0.1846 & 0.4730 \\
    DaVinci-MagiHuman-256p & 15B & 0.6067 & \underline{4.9288} & 6.0619 & 2.4117 & 6.2781 & 0.1311 & 5.7972 & 0.2565 & 0.4150 \\
    \DreamXTableAltRow{}
    DaVinci-MagiHuman-512p & 15B & 0.6013 & {\bfseries 4.9361} & 6.0720 & 2.4156 & 6.2836 & 0.1228 & 6.8997 & 0.2653 & 0.4010 \\
    \midrule
    \DreamXTableAltRow{}
    {\bfseries Ours} & 7B & 0.6568 & 4.7210 & \underline{6.0818} & 2.1739 & 6.3519 & \underline{0.1112} & \underline{7.8018} & 0.2608 & 0.1902 \\
    {\bfseries Ours (RL)} & 7B & 0.6573 & 4.7463 & {\bfseries 6.0929} & 2.4072 & \underline{6.4094} & 0.1232 & {\bfseries 7.8361} & \underline{0.2677} & {\bfseries 0.1351} \\
    \DreamXTableAltRow{}
    {\bfseries Ours (Refiner)} & 7B & {\bfseries 0.6930} &
    4.7463 &
    {\bfseries 6.0929} &
    2.4072 &
    \underline{6.4094} &
    0.1232 & 7.6979 & 0.2675 & \underline{0.1731} \\
    \bottomrule
  \end{tabularx}
  \DreamXTableNote{\raggedright VQ denotes the reported video-quality aggregate.
    CE, CU, PC, and PQ are the AudioBox Aesthetics dimensions Content Enjoyment,
    Content Usefulness, Production Complexity, and Production Quality. DeSync is
    the Synchformer-based audiovisual desynchronization score used in the
    MOVA-style evaluation. PC, WER, and DeSync are lower-is-better under this
    protocol; VQ, CE, CU, PQ, LSE-C, and IB are higher-is-better. Precision,
    duration, and inference settings are retained in model names where supplied.}
\end{table}

\begin{table}[htbp]
  \caption{Preliminary quantitative comparison with the larger open-weight
    LTX-2.3 and MiniMax-H3 systems. Metrics are grouped by evaluation target,
    and arrows indicate the preferred direction for each metric.}
  \label{tab:large-open-model-comparison}
  \centering
  \DreamXCompactTableSetup{}
  \setlength{\tabcolsep}{1.5pt}
  \begin{tabularx}{\linewidth}{@{}>{\raggedright\arraybackslash}p{1.55in}
      >{\centering\arraybackslash}p{0.40in}
      *{9}{>{\centering\arraybackslash}X}@{}}
    \toprule
    \DreamXTableHeaderRow{}
    \rule{0pt}{2.8ex} & &
    \multicolumn{1}{c}{\DreamXTableHead{Video}} &
    \multicolumn{4}{c}{\DreamXTableHead{Audio Aesthetics}} &
    \multicolumn{1}{c}{\DreamXTableHead{ASR}} &
    \multicolumn{1}{c}{\DreamXTableHead{Lip Sync}} &
    \multicolumn{2}{c}{\DreamXTableHead{AV-Align}} \\[-0.5pt]
    \arrayrulecolor{white}\cline{3-11}\arrayrulecolor{black}
    \cellcolor{dreamxprimary}\rule[-0.8ex]{0pt}{3.2ex}\multirow{-2}{*}{\DreamXTableHead{Model}} &
    \cellcolor{dreamxprimary}\multirow{-2}{*}{\DreamXTableHead{Params}} &
    \cellcolor{dreamxprimary}\DreamXTableHead{VQ$\uparrow$} &
    \cellcolor{dreamxprimary}\DreamXTableHead{CE$\uparrow$} &
    \cellcolor{dreamxprimary}\DreamXTableHead{CU$\uparrow$} &
    \cellcolor{dreamxprimary}\DreamXTableHead{PC$\downarrow$} &
    \cellcolor{dreamxprimary}\DreamXTableHead{PQ$\uparrow$} &
    \cellcolor{dreamxprimary}\DreamXTableHead{WER$\downarrow$} &
    \cellcolor{dreamxprimary}\DreamXTableHead{LSE-C$\uparrow$} &
    \cellcolor{dreamxprimary}\DreamXTableHead{IB$\uparrow$} &
    \multicolumn{1}{>{\columncolor{dreamxprimary}[\tabcolsep][0pt]}c@{}}{\DreamXTableHead{DeSync$\downarrow$}} \\
    \midrule
    \DreamXTableAltRow{}
    LTX-2.3 & 22B & 0.6285 & 5.1876 & 6.4033 & 2.6585 & 6.7169 & 0.1113 & 7.6768 & 0.3081 & 0.2412 \\
    MiniMax-H3 (open source) & 33B & 0.6429 & 5.2776 & 6.5847 & 2.4428 & 7.0140 & 0.1619 & 8.7354 & 0.3119 & 0.2708 \\
    \midrule
    \DreamXTableAltRow{}
    {\bfseries Ours} & 7B & 0.6568 & 4.7210 & 6.0818 & 2.1739 & 6.3519 & 0.1112 & 7.8018 & 0.2608 & 0.1902 \\
    {\bfseries Ours (RL)} & 7B & 0.6573 & 4.7463 & 6.0929 & 2.4072 & 6.4094 & 0.1232 & 7.8361 & 0.2677 & 0.1351 \\
    \DreamXTableAltRow{}
    {\bfseries Ours (Refiner)} & 7B & 0.6930 &
    4.7463 &
    6.0929 &
    2.4072 &
    6.4094 &
    0.1232 & 7.6979 & 0.2675 & 0.1731 \\
    \bottomrule
  \end{tabularx}
  \DreamXTableNote{\raggedright VQ denotes the reported video-quality aggregate.
    CE, CU, PC, and PQ are the AudioBox Aesthetics dimensions Content Enjoyment,
    Content Usefulness, Production Complexity, and Production Quality. DeSync is
    the Synchformer-based audiovisual desynchronization score used in the
    MOVA-style evaluation. PC, WER, and DeSync are lower-is-better under this
    protocol; VQ, CE, CU, PQ, LSE-C, and IB are higher-is-better. Precision,
    duration, and inference settings are retained in model names where supplied.}
\end{table}

Compared with the larger LTX-2.3 (22B)
\citep{lightricks2026ltx23,lightricks2026ltx23release} and MiniMax-H3 (33B)
\citep{minimax2026h3}, our 7B model remains behind on several metrics. Relative
to Ours (RL), LTX-2.3 is higher on CE (5.1876 vs. 4.7463), CU (6.4033 vs.
6.0929), PQ (6.7169 vs. 6.4094), and IB (0.3081 vs. 0.2677). MiniMax-H3
further leads on CE (5.2776), CU (6.5847), PQ (7.0140), LSE-C (8.7354), and
IB (0.3119), compared with 4.7463, 6.0929, 6.4094, 7.8361, and 0.2677 for Ours
(RL), respectively. These gaps show that the current 7B model has not reached
across-the-board parity with the two larger open-weight baselines, particularly
in audio aesthetics and cross-modal semantics, and it also remains behind
MiniMax-H3 in lip synchronization. The favorable VQ and DeSync results therefore
represent a trade-off rather than overall superiority; closing the remaining
gap will require stronger audio post-training and cross-modal alignment.

\FloatBarrier

\subsection{Refiner Comparison}

Table~\ref{tab:refiner-comparison} shows that our refiner achieves the best
overall trade-off among the refinement methods. It obtains the highest MUSIQ and
MANIQA scores, 0.7073 and 0.4382, respectively, and a competitive aesthetic score
of 0.4911. In addition to improving perceptual quality, our method also achieves
the best LSE-C and IB among the refinement methods, together
with a low DeSync score of 0.1731. These results indicate that our refiner improves perceptual video quality while
better retaining semantic consistency and audio-visual alignment, leading to a
more balanced refinement performance.


\begin{table}[htbp]
  \caption{Comparison of the baseline generator and video restoration/refinement
    methods. Arrows indicate the preferred direction for each metric.}
  \label{tab:refiner-comparison}
  \centering
  \DreamXResultTableSetup{}
  \setlength{\tabcolsep}{2.2pt}
  \begin{tabularx}{\linewidth}{@{}>{\raggedright\arraybackslash}p{1.55in}
      *{6}{>{\centering\arraybackslash}X}@{}}
    \toprule
    \DreamXTableHeaderRow{}
    \DreamXTableHead{Model} &
    \DreamXTableHead{Aesthetic$\uparrow$} &
    \DreamXTableHead{MUSIQ$\uparrow$} &
    \DreamXTableHead{MANIQA$\uparrow$} &
    \DreamXTableHead{LSE-C$\uparrow$} &
    \DreamXTableHead{IB$\uparrow$} &
    \DreamXTableHead{DeSync$\downarrow$} \\
    \midrule
    \rowcolor{dreamxheadergray!18}
    \textcolor{dreamxheadergray}{Baseline}
      & \textcolor{dreamxheadergray}{0.4444}
      & \textcolor{dreamxheadergray}{0.5895}
      & \textcolor{dreamxheadergray}{0.3006}
      & \textcolor{dreamxheadergray}{7.8361}
      & \textcolor{dreamxheadergray}{0.2677}
      & \textcolor{dreamxheadergray}{0.1351} \\
    FlashVSR
      & \mbox{{\bfseries 0.4940}\,{\tiny\textcolor{green!55!black}{$(+.0496)$}}}
      & \mbox{\underline{0.7000}\,{\tiny\textcolor{green!55!black}{$(+.1105)$}}}
      & \mbox{\underline{0.4180}\,{\tiny\textcolor{green!55!black}{$(+.1174)$}}}
      & \mbox{\underline{7.6772}\,{\tiny\textcolor{red!70!black}{$(-.1589)$}}}
      & \mbox{0.2627\,{\tiny\textcolor{red!70!black}{$(-.0050)$}}}
      & \mbox{{\bfseries 0.1604}\,{\tiny\textcolor{red!70!black}{$(+.0253)$}}} \\
    \DreamXTableAltRow{}
    SeedVR
      & \mbox{0.4672\,{\tiny\textcolor{green!55!black}{$(+.0228)$}}}
      & \mbox{0.6710\,{\tiny\textcolor{green!55!black}{$(+.0815)$}}}
      & \mbox{0.3586\,{\tiny\textcolor{green!55!black}{$(+.0580)$}}}
      & \mbox{7.3456\,{\tiny\textcolor{red!70!black}{$(-.4905)$}}}
      & \mbox{0.2640\,{\tiny\textcolor{red!70!black}{$(-.0037)$}}}
      & \mbox{0.1851\,{\tiny\textcolor{red!70!black}{$(+.0500)$}}} \\
    LTX-2.5 Refiner
      & \mbox{0.4824\,{\tiny\textcolor{green!55!black}{$(+.0380)$}}}
      & \mbox{0.6200\,{\tiny\textcolor{green!55!black}{$(+.0305)$}}}
      & \mbox{0.3377\,{\tiny\textcolor{green!55!black}{$(+.0371)$}}}
      & \mbox{7.6489\,{\tiny\textcolor{red!70!black}{$(-.1872)$}}}
      & \mbox{\underline{0.2670}\,{\tiny\textcolor{red!70!black}{$(-.0007)$}}}
      & \mbox{0.2166\,{\tiny\textcolor{red!70!black}{$(+.0815)$}}} \\
    \DreamXTableAltRow{}
    Ours Refiner
      & \mbox{\underline{0.4911}\,{\tiny\textcolor{green!55!black}{$(+.0467)$}}}
      & \mbox{{\bfseries 0.7073}\,{\tiny\textcolor{green!55!black}{$(+.1178)$}}}
      & \mbox{{\bfseries 0.4382}\,{\tiny\textcolor{green!55!black}{$(+.1376)$}}}
      & \mbox{{\bfseries 7.6979}\,{\tiny\textcolor{red!70!black}{$(-.1382)$}}}
      & \mbox{{\bfseries 0.2675}\,{\tiny\textcolor{red!70!black}{$(-.0002)$}}}
      & \mbox{\underline{0.1731}\,{\tiny\textcolor{red!70!black}{$(+.0380)$}}} \\
    \bottomrule
  \end{tabularx}
  \DreamXTableNote{\raggedright Higher is better for Aesthetic, MUSIQ, MANIQA,
    LSE-C, and IB. Lower is better for DeSync. Parenthesized values report the
    raw difference relative to the Baseline (method minus Baseline); green and
    red indicate improved and degraded performance, respectively. Therefore, a
    positive DeSync difference is red because it indicates worse synchronization.
    Bold and underlined values denote the best and second-best results among the
    restoration/refinement methods, respectively; the gray Baseline row is
    excluded from this ranking.}
\end{table}

\FloatBarrier

\subsection{User Study}
\label{sec:user-study}

To complement automatic metrics, we conduct a blind side-by-side human
preference study. Each trial compares \ours{} with one baseline under the same
prompt and playback setting. The comparison cases are randomly sampled from Verse-Bench. The model identities are anonymized and the
left-right order is randomized. Assessors report whether \ours{} wins, ties, or
loses along four dimensions: text-video alignment (TV-Align), audio-video
alignment (AV-Align), video quality, and audio quality. We report all
percentages from the perspective of \ours{}.

\begin{figure}[H]
  \centering
  \includegraphics[width=\linewidth]{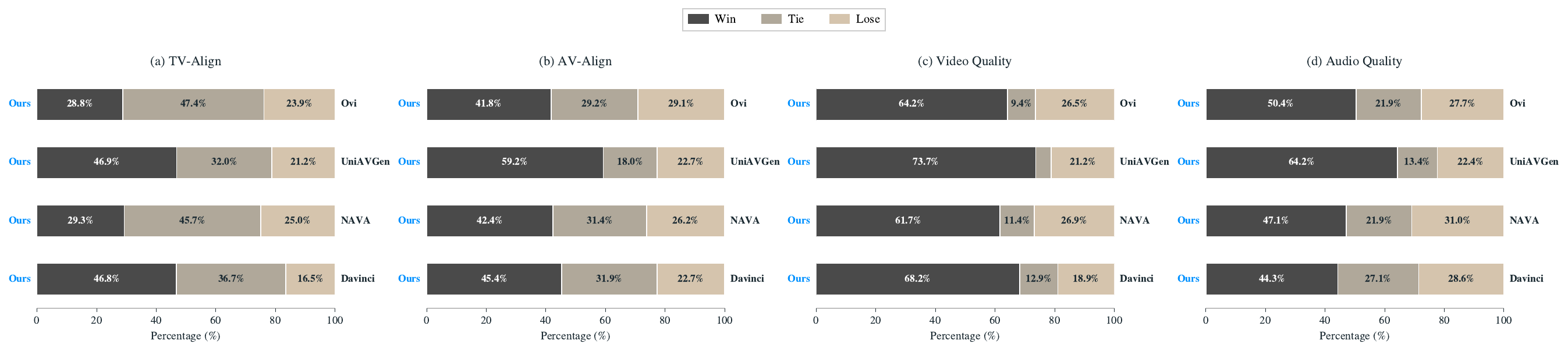}
  \vspace{0.35em}
  \includegraphics[width=\linewidth]{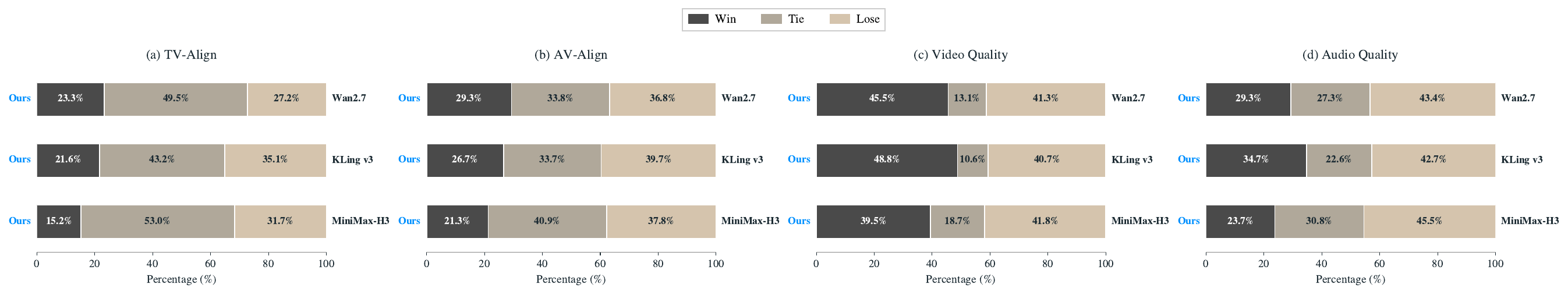}
  \caption{Human preference study comparing \ours{} with Ovi, UniAVGen, NAVA,
    and DaVinci (top), and with Wan2.7, Kling v3, and MiniMax-H3 (bottom).
    Each horizontal stacked bar reports Win/Tie/Lose percentages from the
    perspective of \ours{} under blind side-by-side comparison. }
  \label{fig:user-study-preference}
\end{figure}

As shown in \cref{fig:user-study-preference}, \ours{} records more wins than
losses in every criterion--baseline pair against Ovi, UniAVGen, NAVA, and
DaVinci. The advantage is most pronounced in video quality, with win rates of
64.2\%, 73.7\%, 61.7\%, and 68.2\%, respectively, and remains consistent in
AV-Align, where win rates range from 41.8\% to 59.2\%. TV-Align is closer
against Ovi and NAVA, with tie rates of 47.4\% and 45.7\%, respectively, but
\ours{} still receives more wins than losses in both comparisons.

Against the industrial-scale systems, the observed gap is limited in several
dimensions, particularly video quality: \ours{} wins/loses 45.5\%/41.3\%
against Wan2.7 and 48.8\%/40.7\% against Kling v3, while the comparison with
MiniMax-H3 is close at 39.5\%/18.7\%/41.8\% win/tie/lose. TV-Align also yields
more losses than wins, although tie rates are high (43.2\%--53.0\%), and losses
exceed wins on AV-Align and audio quality against all three systems. These
results should be interpreted within the scope of the sampled subset: highly
dynamic and compositionally complex scenes are under-represented, so the study
does not fully probe the capabilities of these industrial systems.
The findings therefore show that our 7B model remains competitive on the comparatively less challenging cases covered by this study, rather than establishing across-the-board parity; audio fidelity and cross-modal alignment remain the main areas for improvement.

\FloatBarrier

%% file: sections/related_work.tex
\section{Related Work}\label{sec:related-work}

\subsection{Audio-Video Generation Models}

Audiovisual synthesis is often posed directionally---video-to-audio~\citep{luo2023difffoley,zhang2024foleycrafter,cheng2025mmaudio} or audio-to-video for talking-head/human animation~\citep{tian2024emo,lin2025omnihuman,gao2025wans2v,yangmace, yang2026omnidance}---which performs well when one modality can be treated as a fixed timeline, but precludes reciprocal co-generation. Joint diffusion models and diffusion transformers instead sample audio and video together, progressing from coupled denoisers ~\citep{ruan2023mmdiffusion,xing2024seeing} to more integrated backbones and expert compositions~\citep{wang2024avdit,liu2024syncflow,low2025ovi,hacohen2026ltx2,openmoss2026mova,cheng2026unison,wang2025universe,zhang2026uniavgen} and toward shared representations with explicit alignment and physical-time modeling~\citep{ji2026native,liu2025javisdit,liu2026javisditplusplus}. Achieving both semantic correspondence and fine-grained synchronization remains central, motivating physical-time encodings, cross-modal attention, discriminative synchrony objectives~\citep{low2025ovi,wang2025universe,liu2026javisditplusplus,iashin2024synchformer,chung2016outoftime}, and selective fusion mechanisms such as gated attention~\citep{qiu2025gatedattention}. \ours{} follows this trend by keeping modality-specific streams while making bidirectional exchange context dependent.

\subsection{From Single-Modality to Joint Audio-Video Preference Alignment}

Diffusion and flow models~\citep{chu2025uspunifiedselfsupervisedpretraining, yu2026elucidatingsnrtbiasdiffusion, lei2026there,lei2023masked} primarily fit the data distribution and do not directly optimize perceptual quality, instruction following, or human preference, motivating a line of post-training work that aligns generators to explicit rewards. At the image level, methods differ mainly in their optimization paradigm: DDPO casts denoising as a sequential decision process and applies policy-gradient reinforcement learning~\citep{black2023ddpo}, Diffusion-DPO derives a direct preference objective from chosen/rejected pairs without an explicit reward model~\citep{wallace2024diffusiondpo}, and D3PO extends direct preference optimization along the full denoising trajectory to avoid training a separate reward model~\citep{yang2024usinghumanfeedbackfinetune}.

Extending alignment from images to video adds a temporal dimension, so rewards must evaluate not only single-frame aesthetics and text consistency but also temporal coherence, motion quality, subject consistency, and long-range dynamics. VADER backpropagates differentiable rewards through video diffusion~\citep{prabhudesai2024vader}, VideoDPO constructs preference pairs tailored to video generation~\citep{liu2025videodpo}, and reinforcement-learning formulations such as DanceGRPO and Flow-GRPO adapt group-relative policy optimization to diffusion and flow models~\citep{xue2025dancegrpounleashinggrpovisual,liu2025flowgrpotrainingflowmatching}, while DiffusionNFT recasts reward-driven fine-tuning as negative-aware forward-process training~\citep{zheng2026diffusionnftonlinediffusionreinforcement}. These methods are effective for video, yet they still optimize only the visual modality and cannot directly optimize joint audio-video multi-modal tasks.

Joint audio-video generation raises distinct alignment challenges: a single sample simultaneously involves visual quality, audio quality, text consistency, cross-modal semantics, and temporal synchronization; different rewards may conflict; a single global reward cannot clearly attribute improvements to the audio stream, the video stream, or the interaction module; and fine-grained synchronization typically occurs only in local regions such as lip movements, collisions, and sounding objects. JavisDiT++ proposes AV-DPO with chosen/rejected pairs reflecting visual, acoustic, semantic, cross-modal, and synchronization criteria~\citep{liu2026javisditplusplus}, and OmniNFT extends negative-aware forward-process training to the multi-modal setting~\citep{zhang2026omninftmodalitywiseomnidiffusion}. Such audiovisual alignment introduces additional pitfalls: improving one modality can hide regressions in the other, and optimizing a synchrony score can exploit evaluator artifacts---so controlled same-clip temporal shifts and held-out human comparisons remain necessary to distinguish genuine alignment improvements from reward-model shortcutting. Existing work proves the feasibility of audio-video preference optimization, but how to perform stable post-training for a joint generator with modality-specific backbones and a bidirectional interaction module remains open. \ours{} addresses this with (1) a first-frame-conditioned RL setting; (2) decomposed audio, video, and cross-modal feedback that keeps quality, instruction adherence, semantic correspondence, event timing, and lip synchronization as separable, auditable components; and (3) integration with \ours{}'s bidirectional context-aware interaction.

\subsection{Video Refinement and Few-Step Distillation}

Video restoration and super-resolution propagate information across frames for detail and temporal consistency. BasicVSR++ and RealBasicVSR use recurrent propagation for clean and real-world degradations~\citep{chan2022basicvsrpp,chan2022realbasicvsr}, and VRT introduces a recurrent restoration transformer~\citep{liang2024vrt}. Diffusion-based refiners (StableVSR, Upscale-A-Video, VEnhancer) leverage generative priors for realistic enhancement~\citep{rota2024enhancingperceptualqualityvideo,zhou2024upscaleavideo,he2024venhancer}, while large-scale systems such as VideoGigaGAN and SeedVR target more detailed and general restoration~\citep{zhang2025videogigagan,wang2025seedvrseedinginfinitydiffusion}; SeedVR in particular is a 2.48B-parameter diffusion transformer with shifted-window attention that performs generic real-world video restoration at arbitrary length and resolution~\citep{wang2025seedvrseedinginfinitydiffusion}. Recent industrial and interactive systems push this further: LPM scales generative video restoration with progressive training and temporal-pyramid inference for arbitrarily long videos in production~\citep{zeng2026lpm10videobasedcharacter}, SparkVSR lets users super-resolve sparse keyframes and then propagates the high-resolution priors to the full video~\citep{yu2026sparkvsrinteractivevideosuperresolution}, and UltraFlash cascades streaming components for real-time high-resolution generation and super-resolution (roughly 30~FPS at 1K)~\citep{luxury2026ultraflashscalingrealtime}.

Separately, few-step generation uses distillation to reduce sampling cost. DMD and DMD2 match teacher and student distributions, and LADD improves high-resolution efficiency~\citep{yin2024onestep,yin2024dmd2,sauer2024ladd}. These distillation principles have been carried into video super-resolution: DUO-VSR applies distribution-matching distillation with a dual-stream design for one-step inference, achieving roughly a 50$\times$ speed-up over SeedVR-7B~\citep{lv2026duovsrdualstreamdistillationonestep}, FlashVSR is the first diffusion-based one-step streaming VSR, combining three-stage distillation, locality-constrained sparse attention, and a tiny conditional decoder to reach real-time ${\sim}17$~FPS at 768$\times$1408~\citep{zhuang2025flashvsrrealtimediffusionbasedstreaming}, and LiteVSR adds a lightweight adapter (11.25\% trainable parameters) with time-dependent cross-attention aligning intermediate denoising states, compatible with single-step inference~\citep{cao2026litevsrlightweightadaptationfrozen}. For autoregressive video, Self-Forcing and Causal Forcing reduce train--test mismatch by training on the model's own generated context~\citep{huang2025selfforcing,zhu2026causalforcing}. Our refiner operates under a different contract: the full low-resolution clip is available before refinement, enabling bidirectional full-clip attention for both teacher and few-step student. We therefore distill under the offline refinement setting rather than imposing a causal constraint.

%% file: sections/conclusion.tex
\section{Conclusion}\label{sec:conclusion}

We presented \ours{}, an open framework for joint audio-video generation. Its
implemented generator preserves dedicated audio and video streams, aligns them
on a shared timeline, and exchanges information through bidirectional attention
regulated by a hidden- and context-dependent gate. Content-adaptive directional
training and full-model consolidation place A2V, V2A, and joint generation
within one flow-matching framework.
The broader system connects this generator to two next steps: multimodal
reinforcement learning for user-facing alignment and a bidirectional few-step
Refiner targeting 2K output. In this initial report these stages are stated as
designs with explicit validation requirements, not as completed empirical
claims. By releasing the evaluated model variant, inference stack, and
evaluation tools with explicit provenance, our goal is to broaden access to
synchronized audio-video generation and high-resolution generative video
research.

%% file: sections/acknowledgements.tex
\DreamXAuthorAcknowledgements{%
  The contributor list, institutional acknowledgements, and release credits
  will be finalized with the public version of this report.}